\documentclass[letterpaper, 10 pt, conference]{ieeeconf}  

\IEEEoverridecommandlockouts                              

\usepackage{times}
\usepackage{soul}
\usepackage{url}
\usepackage{graphicx}
\usepackage{amsmath}
\usepackage{amssymb}
\usepackage{booktabs}
\usepackage{multirow}
\usepackage{array}
\usepackage{cite}
\usepackage{float}
\usepackage{listings}
\usepackage{xcolor}

\usepackage[linesnumbered]{algorithm2e}
\RestyleAlgo{ruled}
\SetKwComment{Comment}{/* }{ */}
\SetKwInput{KwRequire}{Require}

\usepackage{fancyvrb}

\definecolor{ao(english)}{rgb}{0.0, 0.5, 0.0}

\title{\LARGE \bf Robot Task Planning and Situation Handling in Open Worlds}

\author{Yan Ding$^1$,
    Xiaohan Zhang$^1$,
    Saeid Amiri$^1$,
    Nieqing Cao$^1$,
    Hao Yang$^2$,
    Chad Esselink$^2$,
    Shiqi Zhang$^1$
\thanks{$^1$Yan Ding, Xiaohan Zhang, Saeid Amiri, Nieqing Cao, and Shiqi Zhang are with SUNY Binghamton.
    \texttt{\{yding25, xzhan244, samiri1, ncao1, zhangs\}@binghamton.edu}}%
\thanks{$^{2}$Hao Yang and Chad Esselink are with Ford Motor Company. \texttt{\{hyang1, cesselin\}@ford.com}}%
}

\begin{document}
\maketitle


\begin{abstract}
Automated task planning algorithms have been developed to help robots complete complex tasks that require multiple actions.
Most of those algorithms have been developed for ``closed worlds'' assuming complete world knowledge is provided.
However, the real world is generally open, and the robots frequently encounter unforeseen situations that can potentially break the planner's completeness. 
This paper introduces a novel algorithm (COWP) for open-world task planning and situation handling that dynamically augments the robot's action knowledge with task-oriented common sense.
In particular, common sense is extracted from Large Language Models based on the current task at hand and robot skills.
For systematic evaluations, we collected a dataset that includes 561 execution-time situations in a dining domain, where each situation corresponds to a state instance of a robot being potentially unable to complete a task using a solution that normally works.
Experimental results show that our approach significantly outperforms competitive baselines from the literature in the success rate of service tasks.
Additionally, we have demonstrated COWP using a mobile manipulator.
The project website is available at: \url{https://cowplanning.github.io/}, where \textcolor{blue}{a more detailed version can also be found.} 
This version has been accepted for publication in Autonomous Robots.
\end{abstract}



\section{Introduction}
	
\begin{figure}[h]
\vspace{1em}
\centering
\includegraphics[width=0.98\columnwidth]{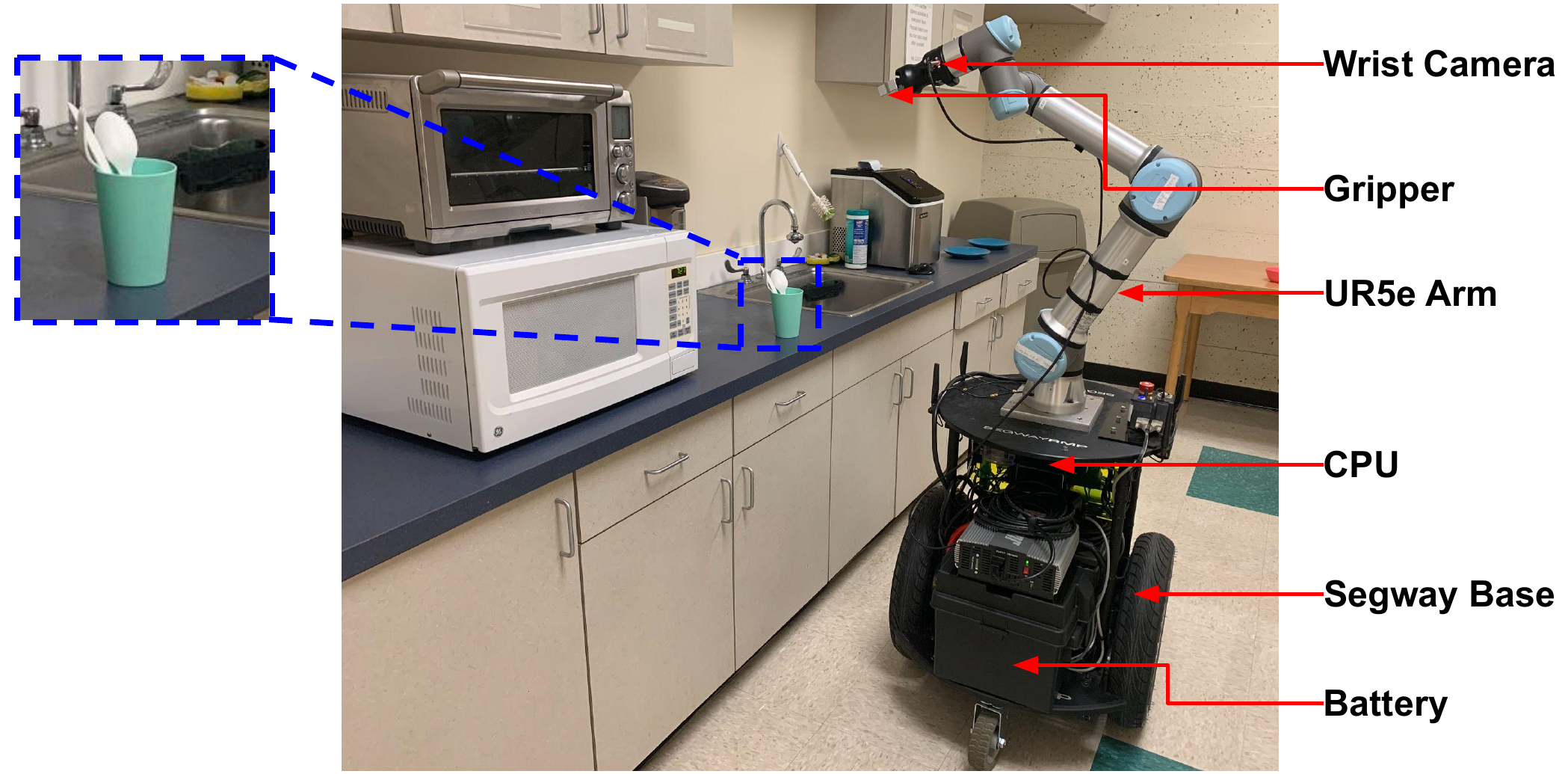}
\caption{An illustrative example of a \textit{situation} in the real world, encountered during the execution of the plan ``delivering a cup to a human for drinking water.'' 
The robot approached a cabinet in a kitchen room on which a cup was located.
The robot then found the cup to be delivered. 
Before grasping it, however, the robot detected a situation that \textit{the cup was occupied with a fork, a knife, and a spoon}. 
This situation prevented the robot from performing the current action (i.e., grasping) and rendered normal solutions for drinking water invalid. 
A mobile service robot with a UR5e arm on a Segway RMP-110 base was used for demonstrations in this work.
}\label{fig:mobile_manipulator}
\end{figure}

Robots that operate in the real world frequently encounter complex tasks that require multiple actions.
Automated task planning algorithms have been developed to help robots sequence actions to accomplish those tasks~\cite{ghallab2016automated}.
Closed world assumption (CWA) is a presumption that was developed by the knowledge representation community and states that ``statements that are true are also known to be true''~\cite{reiter1981closed}. 
Most current task planners have been developed for closed worlds, assuming complete domain knowledge is provided and one can enumerate all possible world states~\cite{knoblock1991characterizing, hoffmann2001ff, nau2003shop2, helmert2006fast}.
However, the real world is ``open'' by nature, and unforeseen situations are common in practice~\cite{hanheide2017robot}. 
As a consequence, current automated task planners tend to be fragile in open worlds rife with situations.
Fig.~\ref{fig:mobile_manipulator} shows an example situation:  \textit{Aiming to grasp a cup for drinking water, a robot found the cup was occupied with forks and knives}. 
Although one can name many such situations, it is impossible to provide a complete list of them. 
To this end, researchers have developed open-world planning methods for robust task completions in real-world scenarios~\cite{jiang2019open, chernova2020situated, hanheide2017robot, kant2022housekeep, huang2022language, ahn2022can}.


There are mainly two ways of performing open-world planning without humans in the loop in the literature.
One relies on dynamically building an external knowledge base to assist a pre-defined task planner, where this  knowledge base is usually constructed in an automatic way using external information~\cite{jiang2019open,chernova2020situated,hanheide2017robot}. 
Such external knowledge bases are considered bounded due to their representation and knowledge source, which limits the ``openness'' of their task planners. 
Another (more recent) way of building open-world planners is to leverage Large Language Models (LLMs)~\cite{bommasani2021opportunities}. 
LLMs have significantly improved the performance of downstream language tasks in recent years~\cite{brown2020language, zhang2022opt, vaswani2017attention, radford2019language}. 
Recent research has demonstrated that those LLMs contain a wealth of commonsense knowledge~\cite{wang2021can, li2022pre, ahn2022can, west2021symbolic}. 
While it is an intuitive idea of extracting common sense from LLMs for task planning~\cite{ahn2022can, huang2022language, elsweiler2022food, kant2022housekeep}, there is a fundamental challenge for robots to ``ground'' \emph{domain-independent} commonsense knowledge~\cite{davis2015commonsense} to specific domains that are featured with many \emph{domain-dependent} constraints. 
We propose to acquire common sense from LLMs, and aim to improve the task completion and situation handling skills of service robots. 

In this paper, we develop a robot task planning framework, called \textit{Common sense-based Open-World Planning}~(\textbf{COWP}), that uses an LLM (GPT-3 in our case~\cite{brown2020language}) for dynamically augmenting automated task planners with external task-oriented common sense.
COWP is based on classical planning and leverages LLMs to augment action knowledge (action preconditions and effects) for task planning and situation handling. 
The \textbf{main contribution} of this work is a novel integration of a pre-trained LLM with a knowledge-based task planner. 
Inheriting the desirable features from both sides, COWP is well grounded in specific domains while embracing commonsense solutions at large. 

For systematic evaluations, we have created a dataset with 561 execution-time situations collected from a \textit{dining domain}~\cite{garrett2020online, gordon2021leveraging, puig2018virtualhome} using a crowd-sourcing platform, where each situation corresponds to an instance of a robot not being able to perform a plan (that normally works). 
According to experimental results, we see COWP performed significantly better than three literature-selected baselines~\cite{helmert2006fast, jiang2019open, huang2022language} in success rate. 
We implemented and demonstrated COWP using a mobile manipulator. 


\section{Background and Related Work}
In this section, we first briefly discuss classical task planning methods that are mostly developed under the closed world assumption. 
We then summarize three families of open-world task planning methods for robots, which are grouped based on how unforeseen situations are addressed. 

\vspace{.5em}
\noindent
\textbf{Classical Task Planning for Closed Worlds:}
Closed world assumption (CWA) indicates that an agent is provided with complete domain knowledge, and that all statements that are true are known to be true by the agent~\cite{reiter1981closed}. 
Most automated task planners have been developed under CWA~\cite{ghallab2016automated,haslum2019introduction,nau2003shop2}. 
Although robots face the real world that is open by nature, their planning systems are frequently constructed under the CWA~\cite{hanheide2017robot,jiang2019task,galindo2008robot}. 
The consequence is that those robot planning systems are not robust to unforeseen situations at execution time. 
In this paper, we aim to develop a task planner that is aware of and able to handle unforeseen situations in open-world scenarios.

\vspace{.5em}
\noindent
\textbf{Open-World Task Planning with Human in the Loop:} 
Task planning systems have been developed to acquire knowledge via human-robot interaction to handle open-world situations~\cite{perera2015learning, amiri2019augmenting, tucker2020learning}. 
For instance, researchers created a planning system that uses dialog systems to augment their knowledge bases~\cite{perera2015learning}, whereas Amiri et al. (2019) further modeled the noise in language understanding~\cite{amiri2019augmenting}.
Tucker et al. (2020) enabled a mobile robot to ground new concepts using visual-linguistic observations, e.g., to ground the new word ``box'' given command of ``move to the box'' by exploring the environment and hypothesizing potential new objects from natural language~\cite{tucker2020learning}. 
The major difference from those open-world planning methods is that COWP does not require human involvement.

\vspace{.5em}
\noindent
\textbf{Open-World Task Planning with External Knowledge:}
Some existing planning systems address unforeseen situations by dynamically constructing an external knowledge base for open-world reasoning.
For instance, researchers have developed object-centric planning algorithms that maintain a database about objects and introduce new object concepts and their properties (e.g., location) into their task planners~\cite{jiang2019open, chernova2020situated}.
For example, Jiang~et~al.~(2019) developed an object-centric, open-world planning system that dynamically introduces new object concepts through augmenting a local knowledge base with external information~\cite{jiang2019open}.
In the work of Hanheide et al. (2017), additional action effects and assumptive actions were modeled as an external knowledge to explain the failure of task completion and compute plans in open worlds~\cite{hanheide2017robot}.
A major difference from their methods is that COWP employs an LLM that is capable of responding to any situation, whereas the external knowledge sources of those methods limits the openness of their systems. 

\vspace{.5em}
\noindent
\textbf{Open-World Task Planning with LLMs:}
LLMs can encode a large amount of common sense from corpus~\cite{liu2021pre} and have been applied to robot systems to complete high-level tasks. 
Example LLMs include BERT~\cite{devlin2018bert}, GPT-2~\cite{radford2019language}, GPT-3~\cite{brown2020language}, and OPT~\cite{zhang2022opt}. 
For example, Kant et al. (2022) developed a household robot that uses a fine-tuned LLM to reason about rearrangements of new objects~\cite{kant2022housekeep}. 
Other teams used LLMs to compute plans for high-level tasks specified in natural language (e.g., ``make breakfast'') by sequencing actions~\cite{huang2022language, ahn2022can, huang2022inner}. 
Different from the above-mentioned approaches, in addition to commonsense knowledge extracted from LLMs, our system utilizes rule-based action knowledge from human experts. 
As a result, our planning system can be better grounded to specific domains, and is able to incorporate common sense to augment robot capabilities supported by predefined skills. 
\begin{figure*}[h]
\vspace{.5em}
\centering
\includegraphics[width=2.0\columnwidth]{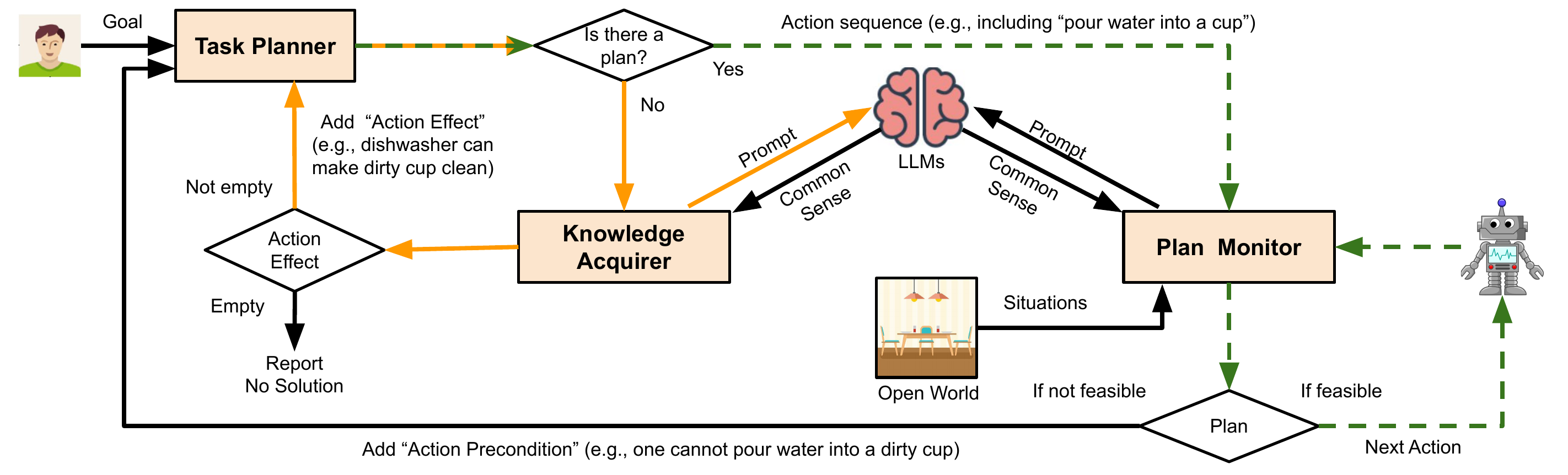}
\caption{An overview of \textbf{COWP} that includes the three key components of Task Planner (provided as prior knowledge under closed-world assumption), Knowledge Acquirer, and Plan Monitor. 
The \textbf{\color{ao(english)} green} (dashed) loop represents a plan execution process where the robot does encounter no situation, or these situations have no impact on the robot's plan execution. 
The \textbf{\color{orange} orange} loop is activated when the robot's current (closed-world) task planner is unable to develop a plan, which activates Knowledge Acquirer to augment the task planner with additional action effects utilizing common sense. }
\label{fig:overview}
\end{figure*}
\vspace{-0.5em}

\section{Algorithm}
\label{sec:alg}
In this section, we first provide a problem statement and then present our open-world planning approach called Common sense-based Open-World Planning (COWP). 

\subsection{Problem Description}
A classical task planning problem is defined by a domain description and a problem description. 
The domain description includes a set of actions, and  action preconditions and effects. 
The problem description includes an initial state and goal conditions. 
In this paper, such a classical planning system is referred to as a \textbf{closed-world task planner}. 
In our case, we provide the robot with a predefined closed-world task planner (implemented using PDDL~\cite{aeronautiques1998pddl}), and an LLM (GPT-3 in our case). 
PDDL, an action-centered language, is designed to formalize Artificial Intelligence (AI) planning problems, allowing for a more direct comparison of planning algorithms and implementations~\cite{aeronautiques1998pddl}.
A \textbf{situation} is defined as an unforeseen world state that potentially prevents an agent from completing a task using a solution that normally works. 
The \textbf{goal} of an open-world planner is to compute plans and handle situations towards completing service tasks or reporting ``no solution'' as appropriate.



\begin{algorithm}
\small
\caption{COWP algorithm}\label{alg:COWP}
\KwRequire{Task Planner, Plan Monitor, Knowledge Acquirer, and an LLM}
\KwIn{A domain description, and a problem description}
\While{\rm task is not completed}
{
    Compute a plan \textit{pln} using Task Planner given the domain description and the problem description\label{l:tp}\;
    \For{\rm each action in \textit{pln}\label{l:for_s}}
    {   
        Evaluate the feasibility of the current plan using Plan Monitor given a situation\label{l:pm_e}\;
        \eIf{\rm a plan is not feasible\label{l:sit}}
        {
                Add an action precondition to Task Planner\label{l:ka_s}\;
                Compute a new task plan $\textit{pln}'$, and \textit{pln} $\leftarrow$ $\textit{pln}'$\label{l:tp'}\;
                \If{\em $\textit{pln}$ is empty\label{l:pln'_empty}}
                {
                    Extract common sense (action effects) using Knowledge Acquirer\label{l:ka}\;
                    \eIf{\em action effect is not empty}
                    {
                        Add an action effect to Task Planner\label{l:ka_effect}\;
                        Compute a new task plan $\textit{pln}''$, and \textit{pln} $\leftarrow$ $\textit{pln}''$\label{l:tp''}\;
                    }
                    {
                        Report ``no solution''\label{l:quit}\;
                    }
                }\label{l:ka_e}
        }
        {
            Execute the next action by the robot\label{l:exe_act}\;
        }
    }\label{l:for_e}
}
\end{algorithm}

\subsection{Algorithm Description}
Fig.~\ref{fig:overview} illustrates the three major components (yellow boxes) of our COWP framework.
\textbf{Task Planner} is used for computing a plan under the closed-world assumption and is provided as prior knowledge in this work. 
\textbf{Plan Monitor} evaluates the overall feasibility of the current plan using common sense. 
\textbf{Knowledge Acquirer} is for acquiring common sense to augment the robot's action effects when the task planner generates no plan.

Algorithm~\ref{alg:COWP} describes how the components of COWP interact with each other. 
Initially, Task Planner generates a satisficing plan based on the goal provided by a human user in \textbf{Line~\ref{l:tp}}. 
After that, the actions in the plan are performed sequentially by a robot in the for-loop of \textbf{Lines~\ref{l:for_s}-\ref{l:for_e}}.
If the current plan remains feasible, as evaluated by Plan Monitor, the next action will be directly passed to the robot for execution in \textbf{Line~\ref{l:exe_act}}; 
otherwise, an action precondition will be added to Task Planner in \textbf{Line~\ref{l:ka_s}}. 
For example, when Task Planner does not know anything about ``dirty cups,'' it might wrongly believe that one can use a dirty cup as a container for drinking water. 
In this situation, \textbf{Line~\ref{l:ka_s}} will add a new statement ``\emph{The precondition of filling a cup with water is that the cup is not dirty}''  into the task planner. 
After the domain description of Task Planner is updated (adding action preconditions or effects), the planner tries to  generate a new plan $\textit{pln}'$ in \textbf{Line~\ref{l:tp'}}. 
If no plan is generated by Task Planner (\textbf{Line~\ref{l:pln'_empty}}), Knowledge Acquirer will be activated for knowledge augmentation with external task-oriented common sense in \textbf{Line~\ref{l:ka}}. 
If the extracted common sense includes action effects, such information will be added into the task planner in \textbf{Line~\ref{l:ka_effect}}. 
For instance, the robot might learn that a chopping board can be used for holding steak, as an action effect that was unknown before. 
This process continues until no additional action effects can be generated (in this case, COWP reports ``no solution'' in \textbf{Line~\ref{l:quit}}) or a plan is generated. 

COWP leverages common sense to augment a knowledge-based task planner to address situations towards robust task completion in open worlds. 
The implementations of \textbf{Lines~\ref{l:pm_e}} and \textbf{\ref{l:ka}} using common sense are non-trivial in practice. 
Next, we describe how commonsense knowledge is extracted from an LLM (GPT-3 in our case) for those purposes.

\subsection{Plan Monitor and Knowledge Acquirer}\label{sec:pm_ka}

\textbf{Plan Monitor} is for evaluating whether there exists a situation that prevents a robot from completing its current task using a solution at hand. 
In particular, the input of the plan monitor includes a set of logical facts collected from the real world. 
The realization of our plan monitor relies on repeatedly querying GPT-3 for each action using the following prompt.
\begin{flushleft}
    Template 1: \texttt{Is it suitable to/that [PERFORM ACTION], if there exists [SITUATION].} 
\end{flushleft}

One example of using the above template is ``\textit{Is it suitable that a robot pours water into a cup, if the cup is broken?}'' 
If any action in the current plan is found infeasible, the whole plan is considered infeasible. 

The objective of \textbf{Knowledge Acquirer} is to acquire common sense for augmenting the task planner's action knowledge for situation handling. 
The following template is for querying an LLM for acquiring common sense about action effects.
\begin{flushleft}
    Template 2: \texttt{Is it suitable to/that [PERFORM ACTION] on/in/with/at/over [OBJECT]?}
\end{flushleft}  
where an example of using Template 2 is ``\emph{Is it suitable to hold steak with a chopping board?}''  
As a result, an additional action effect that ``If a robot performs an action of unloading steak onto a chopping board, the steak will be held on the chopping board'' will be added to the task planner. 

Continuing the ``chopping board'' example, additional action effects introduced through Template 2 might enable the task planner to generate many feasible plans, e.g., using a plate (pen, or chopping board) to hold steak. 
It can be difficult for the task planner to evaluate which plan makes the best sense. 
To this end, we develop the following template (Template 3) for selecting a task plan of the highest quality amon/g those satisficing plans: 


\begin{flushleft}
    Template 3: \texttt{There are some objects, such as  [OBJ-1, OBJ-2, ..., and OBJ-N]. Which is the most suitable for [CURRENT TASK]?}
\end{flushleft}
where in the ``chopping board'' example, we expect the LLM to respond by suggesting ``plate'' being the most suitable for holding steak among those mentioned items. 

\begin{figure*}
\centering
\includegraphics[width=1.95\columnwidth]{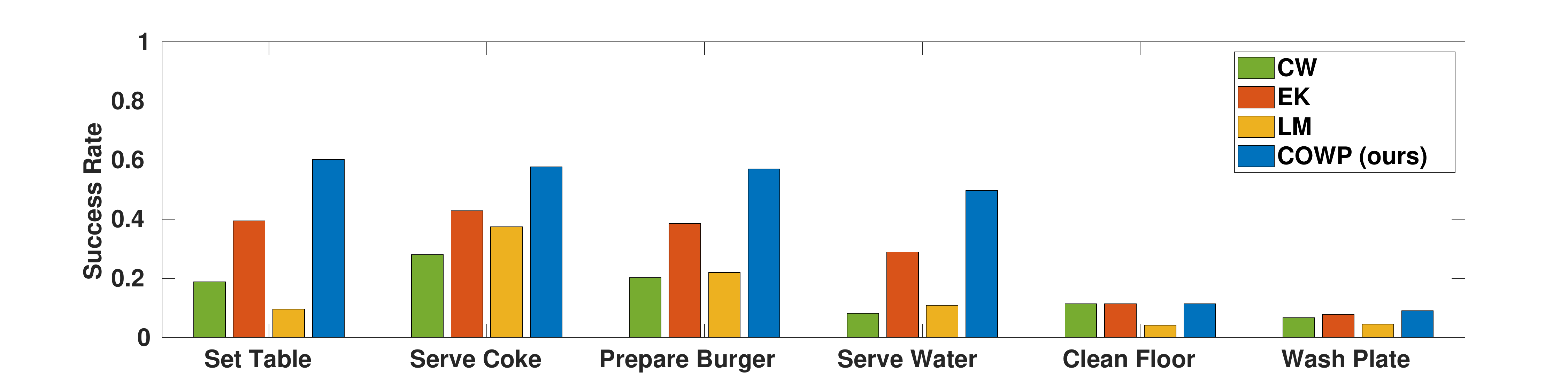}
\vspace{-0.8em}
\caption{Overall performances of COWP (ours) and three baseline methods under \textbf{six different tasks}, where the \textit{x-axis} represents the task.
Each success rate value is an average of 150 trials. 
The tasks are ranked based on the performance of COWP, where the very left corresponds to its best performance. 
}\label{fig:COWP_vs_baseline}
\end{figure*}

\begin{figure*}
\vspace{-1.5em}
\centering
\includegraphics[width=1.85\columnwidth]{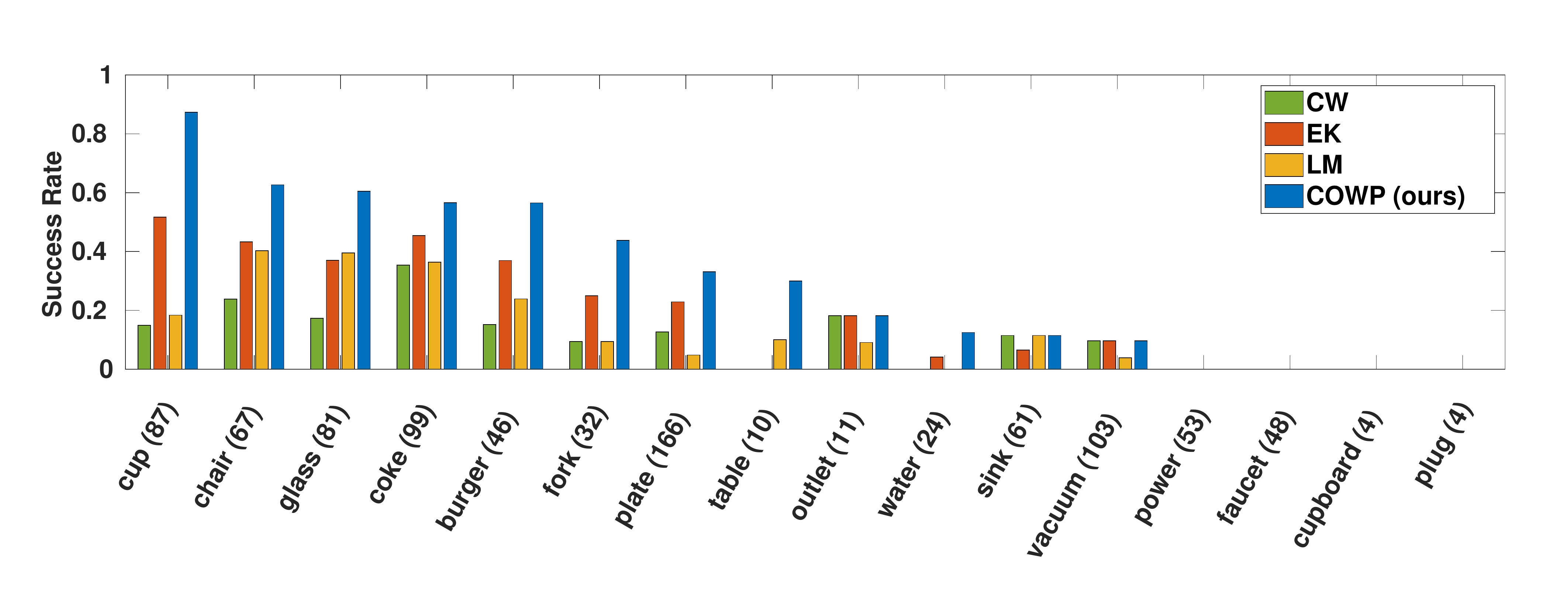}
\vspace{-1.8em}
\caption{Overall performances of COWP (ours) and three baseline methods under \textbf{different objects}, where the \textit{x-axis} represents the object involved in the situation, the number beside each object is the occurrence of the object in our situation dataset, and the \textit{y-axis} represents the success rate.
The objects are ranked based on the performance of COWP, where the very left corresponds to its best performance. 
}\label{fig:objects}
\end{figure*}

\section{Experiments}
In this section, we evaluate the performance of COWP in planning and situation handling tasks.

\vspace{.5em}
\noindent
\textbf{Experimental Setup:} Our experiments were performed in a \textit{dining domain}, where a service robot is tasked with fulfilling a user's service requests in a home setting. 
For simulating dining domains, we extracted six everyday tasks (e.g., serving water, and setting a dining table) and 86 objects (e.g., cup, burger, fork, table, and chair) from an existing dataset~\cite{huang2022language}.  
To create ``situations,'' we randomly selected and spawned half of the available objects in each trial. 
A situation occurs in only one of the steps when the robot executes a task plan in each trial. 
In simulation experiments, we assume that our robot is provided with a perception module for converting raw sensory data into logical facts (e.g., objects and their properties), while the robot still needs to reason about the facts for planning and situation handling.




OpenAI provides several GPT-3 engines with different trade-offs between price and capability, and we used ``\textit{text-davinci-002}'' -- the most capable (expensive) one. 
We selected a ``\textit{temperature}'' of $0.0$ for GPT-3, which minimizes the randomness in the responses. 
Another important parameter is ``\textit{top\_p}'' for diversifying the responses, where we selected 1.0 to maximize response diversity. 
There are a few other parameters, such as ``\textit{maximum length}'', ``\textit{presence\_penalty}'', and ``\textit{frequency\_penalty},'' where we selected 32.0, 0.0, and 0.0, respectively.

\vspace{.5em}
\noindent
\textbf{Situation Dataset for Evaluation:}
To evaluate the performance of COWP in dealing with situations in a dining domain, 
we collected a dataset of execution-time situations using Amazon Mechanical Turk.
Each instance of the dataset corresponds to a situation that prevents a service robot from completing a task. 
We recruited MTurkers with a minimum HIT score of 70. 
Each MTurker was provided with a task description, including steps for completing the task. 
The MTurkers were asked to respond to a questionnaire by identifying one step in the provided plan and describing a situation that might occur in that step. 
For instance, we provided a plan of \emph{1) Walk to dining room, 2) Find a sink, 3) Find a faucet, 4) Turn on faucet, 5) Find a cup...}, which was associated to the task of \emph{Drinking water}. 
Two common responses from the MTurkers were ``\emph{Faucet has no water}'' for Step 4 and ``\emph{Cup is broken}'' for Step 5. 
We received 690 responses from MTurkers, and 561 of them were valid, where the responses were evaluated through both a validation question and a manual filtering process. 
For instance, some MTurkers copied and pasted text that was completely irrelevant, and those responses were removed manually. 
There are at least 92 situations collected for each of the six tasks. 
Note that some situations are similar enough to be grouped together, e.g., ``cup is broken'' and ``cup is shattered'' are considered indistinguishable.
As a result, there are 16-27 distinguishable situations for each task. 
We constructed a situation simulator by randomly sampling indistinguishable situations from our collected dataset for the evaluation purpose in this paper. 
\vspace{.5em}
\noindent
\textbf{Baselines and Evaluation Metrics}:
The evaluation of open-world planners is based on the \textit{success rate} of a robot completing service tasks in a dining domain.
The following three baselines have been used in our experiments: 
\begin{itemize}
    \item Closed World (CW) corresponds to classical task planning developed for closed-world scenarios. 
    In practice, CW was implemented by repeatedly activating the closed-world task planner and updating the current world state after executing each action. 
    \vspace{.5em}
    \item External Knowledge (EK)~\cite{jiang2019open} is a baseline approach that enables a closed-world task planner to acquire knowledge from an external source. 
    In our implementation, this external source provides information about a half of the domain objects. 
    \vspace{.5em}
    \item Language Model (LM)~\cite{huang2022language} is a baseline that leverages GPT-3 to compute task plans, where domain-specific action knowledge is not utilized. 
    The authentic system of the LM baseline~\cite{huang2022language} was not grounded. 
    As a result, their generated plans frequently involve objects unavailable or inapplicable in the current environment. 
    To make a meaningful comparison, we call LM for $N \! \leq \!100$ times until it generated a plan that involves only those objects that are interactable by the robot. 
\end{itemize}

\vspace{.5em}
\noindent
\textbf{Success Criteria:}
Not all ``solutions'' from an open-world task planner are considered correct. 
For instance, GPT-3 suggests that one can use a pan for drinking water, which is technically possible, but uncommon in our everyday life. 
LLM-based task planners might wrongly believe that a good-quality plan is generated, while it is actually not the case. 
To compare with the ground truth, we recruited a second group of people, including six volunteers, to evaluate the performance of different open-world task planners. 
The volunteers included two females and four males aged 20-40, and all were graduate students in engineering fields. 

\vspace{.5em}
\noindent
\textbf{COWP vs. Three Baselines by Task:}
Fig.~\ref{fig:COWP_vs_baseline} shows the results of comparing COWP and three baselines under six tasks in success rate.
In all six tasks, COWP outperformed the three baselines. 
CW and LM produced the lowest success rates in most tasks. 
We believe the poor performance of CW is caused by the incapability of leveraging external information for handling unforeseen situations. 
For LM, its poor performance is caused by its weakness in grounding general commonsense knowledge in specific domains. 
As a result, many ``solutions'' generated by LM are not executable by the robot. 

It is quite interesting to see that open-world planners (including COWP) work better in some tasks than the others. 
For instance, in the ``set table'' task, there are many ``missing object'' situations, and it happened that the robot could easily find alternative tableware objects to address those situations under the assistance of GPT-3. 
Some tasks such as ``Clean Floor'' are more difficult because many situations are beyond the robot's capabilities, e.g., power outage and broken plugs.

\begin{figure*}[t]
\centering
\includegraphics[width=1.9\columnwidth]{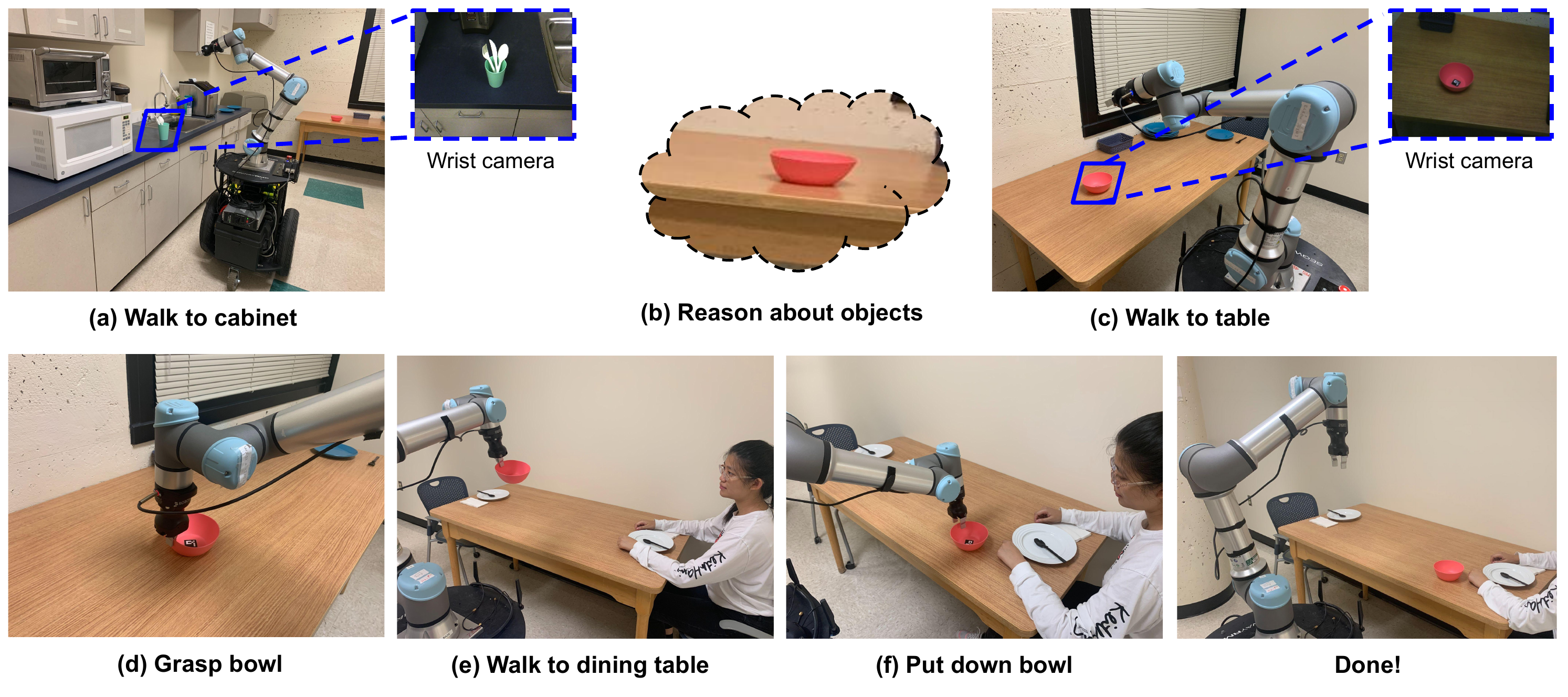}
\vspace{-0.5em}
\caption{
An illustrative example of COWP for open-world planning, where the robot was tasked with ``delivering a cup for drinking water.'' 
\textbf{(a)} The robot walked to a cabinet, and located a cup on the cabinet. 
However, the robot found a situation that there were objects in the cup (a knife, a fork, and a spoon in this case). 
This observation was entered into the plan monitor, which queried GPT-3, and suggested that the planned action ``grasp'' was not applicable given the occupied cup. 
Accordingly, COWP updated its task planner by adding the new information that one cannot pour water into a non-empty cup. 
\textbf{(b)}~The robot reasoned about other objects that were available in the environment, and queried GPT-3 to update the task planner about whether those objects can be used for drinking water -- details in Section~\ref{sec:alg}. 
It happened that the robot learned a bowl could be used for drinking water. 
\textbf{(c)} A new plan of delivering a bowl to the human for drinking water was generated. 
Following the new plan, the robot walked to the table on which a bowl was located. 
\textbf{(d)} The robot grasped the bowl after observing it using vision. 
\textbf{(e)} The robot navigated to the dining table with the bowl. 
\textbf{(f)} The robot put down the bowl onto the dining table, and explained that a bowl was served due to the cup being occupied, which concluded the planning and execution processes. 
}\label{fig:real_demo}
\end{figure*}

\vspace{.5em}
\noindent
\textbf{COWP vs. Three Baselines by Object:}
Fig.~\ref{fig:objects} compares the performance of COWP and three baselines in success rate under \textit{different objects}.
The \textit{x-axis} represents the objects in our situation dataset and their occurrences, and the \textit{y-axis} represents the performance of four planning methods. 
For instance, common situations about cup include ``dirty cup,'' ``broken cup,'' and ``missing cup.'' 
COWP (ours) performed the best over all objects, while some baselines produced comparable performances over some objects. 
An important observation is that COWP produced higher success rates in situations involving ``cup,'' ``coke,'' ``glass,'' and ``chair.'' 
This is because many of those relevant situations are about missing objects, and the robot can easily find their alternatives in dining domains. 
By comparison, those situations involving ``power'', ``faucet'', ``cupboard'', and ``plug'' are more difficult to the four methods (including COWP), because addressing those situations frequently require skills beyond the robot's capabilities. 





\vspace{.5em}
\noindent
\textbf{Robot Demonstration: }
We demonstrated COWP using a mobile service robot that was tasked with delivering a cup for drinking water. 
Our robot is equipped with a UR5e arm, and is based on Segway RMP-110. 
The robot uses a Robotiq Wrist Camera for object detection, and is capable of performing basic navigation and manipulation behaviors. 
For visual scene analysis, the robot detects objects, and then based on their geometric relationships estimates spatial relationships between objects. 
While there exist more powerful tools for visual scene analysis, computer vision is not our focus, and our goal here is to implement a basic perception component to close the perceive-reason-act loop.  



Fig.~\ref{fig:real_demo} shows a real-world demonstration where a robot used COWP for planning to complete the service task of ``Deliver a cup for drinking water.'' 
The initial plan included the following actions for the robot: 
\begin{enumerate}
    \item Walk to a cabinet on which a cup is located, 
    \item Grasp the cup after locating it, 
    \item Walk to dining table, and 
    \item Put down the cup to a table where the human is seated. 
\end{enumerate}

However, a situation was observed after executing Action~\#1, and the robot found \textit{the cup is occupied}. 
The robot initially did not know whether this affects its plan execution. 
After querying GPT-3, the robot learned that one cannot pour water into an occupied cup, which renders its current plan infeasible. 
COWP enabled the robot to reason about other objects of the environment. 
The robot iteratively queried GPT-3 about whether $X$ can be used for drinking water (using Prompt Template 2 in Section~\ref{sec:alg}), where $X$ is one of the objects from the environment. 
It happened that the robot learned ``bowl'' can be used for drinking water. 
With this newly learned, task-oriented commonsense knowledge, COWP successfully helped the robot generate a new plan, which used the bowl instead, to fulfill the service request. 

We have generated a demo video that has been uploaded as part of the supplementary materials.

\section{Conclusion and Future Work}\label{sec:conclusion}

In this paper, we develop a Large Language Model-based open-world task planning system for robots, called COWP, towards robust task planning and situation handling in open worlds. 
The novelty of COWP points to the integration of a classical, knowledge-based task planning system, and a pretrained language model for commonsense knowledge acquisition. 
The marriage of the two enables COWP to ground domain-independent commonsense knowledge to specific task planning problems. 
To evaluate COWP systematically, we collected a situation dataset that includes 561 execution-time situations in a dining domain. 
Experimental results suggest that COWP performed better than existing task planners developed for closed-world and open-world scenarios. 
We also provided a demonstration of COWP using a mobile manipulator working on delivery tasks, which provides a reference to COWP practitioners for real-world applications. 

It is evident that prompt design significantly affects the performance of Large Language Models (LLMs)~\cite{zhang2021differentiable}, which has motivated the recent research on prompt engineering. 
For instance, we tried replacing ``suitable'' with ``possible'' and ``recommended,'' in the prompt templates and observed a decline in the system performance. 
Researchers can improve the prompt design of COWP in future work. 
There is the potential that other LLMs (other than GPT-3) can produce better performance in open-world planning, and their performances might be domain-dependent, which can lead to very interesting future research. 
We present a complete implementation of COWP on a real robot, while acknowledging that there are many ways to improve the implementations. 
For instance, one can use a more advanced visual scene analysis tool to generate more informative observations for situation detection, or equip the robot with more skills (such as wiping a table,  moving a chair, and opening a door) to deal with situations that cannot be handled now. 

\bibliographystyle{IEEEtran}
\bibliography{ref}  

\end{document}